\documentclass[runningheads]{llncs}
\usepackage{graphicx}
\usepackage{pgfplots}  
\usepackage{multirow} 
\pgfplotsset{compat=1.17}
\begin{document}
\title{Context Matters: A Strategy to Pre-train Language Model for Science Education}
\titlerunning{Context Matters: A Strategy to Pre-train Language Model for Science Education}



\author{Zhengliang Liu\inst{1*}
\and Xinyu He \inst{1*}
\and Lei Liu\inst{2}
\and Tianming Liu \inst{1**}
\and Xiaoming Zhai\inst{1**}}


\institute{University of Georgia, Athens, GA 30666, USA 
\\
\and Educational Testing Service, Princeton, NJ, USA\\
** Co-First Author\\
*** Corresponding Authors:\\ 
\email{tliu@uga.edu}\\
\email{xiaoming.zhai@uga.edu}
}

\maketitle              
\begin{abstract}
This study aims at improving the performance of scoring student responses in science education automatically. BERT-based language models have shown significant superiority over traditional NLP models in various language-related tasks. However, science writing of students, including argumentation and explanation, is domain-specific. In addition, the language used by students is different from the language in journals and Wikipedia, which are training sources of BERT and its existing variants. All these suggest that a domain-specific model pre-trained using science education data may improve model performance. However, the ideal type of data to contextualize pre-trained language model and improve the performance in automatically scoring student written responses remains unclear. Therefore, we employ different data in this study to contextualize both BERT and SciBERT models and compare their performance on automatic scoring of assessment tasks for scientific argumentation. We use three datasets to pre-train the model: 1) journal articles in science education, 2) a large dataset of students' written responses (sample size over 50,000), and 3) a small dataset of students' written responses of scientific argumentation tasks. Our experimental results show that in-domain training corpora constructed from science questions and responses improve language model performance on a wide variety of downstream tasks. Our study confirms the effectiveness of continual pre-training on domain-specific data in the education domain and demonstrates a generalizable strategy for automating science education tasks with high accuracy. We plan to release our data and SciEdBERT models for public use and community engagement. 

\end{abstract}
\section{Introduction}

Writing is critical in science learning because it is the medium for students to express their thought processes. In classroom settings, educators have engaged students in writing explanations of phenomena, design solutions, arguments, etc. \cite{novak2009helping}\cite{zhai2022assessing}, with which students develop scientific knowledge and competence. However, it is time-consuming for teachers to review and evaluate natural language writing, thus preventing the timely understanding of students' thought processes and academic progress. Recent development in machine learning (ML), especially natural language processing (NLP), has proved to be a promising approach to promoting the use of writing in science teaching and learning \cite{zhai2020applying}. For example, various NLP methods have been employed in science assessment practices that involve constructed responses, essays, simulation, or educational games \cite{zhai2020substitution}. In this rapidly developing domain, the state-of-the-art Bidirectional Encoder Representations from Transformers (BERT) model \cite{devlin2018bert}, a transformer-based machine learning architecture developed by Google, demonstrates superiority over other machine learning methods in scoring student responses to science assessment tasks \cite{amerman-a}.

\begin{figure}[ht]
\includegraphics[width=\textwidth]{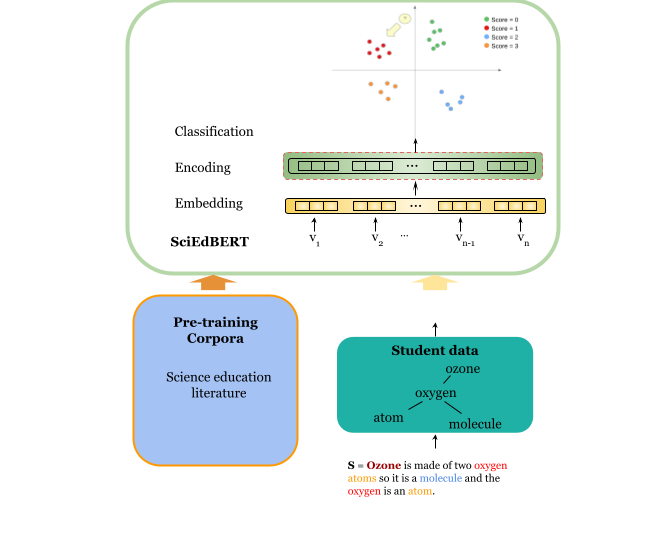}
\caption{The SciEdBERT framework. A student response instance is classified based on the latent representation of word vectors.} \label{fig1}
\end{figure}

Studies have shown that the performance on NLP tasks can be improved by using domain-specific data to contextualize language models \cite{gu2021domain}. Several BERT-based language models, such as SciBERT \cite{beltagy2019scibert}, AgriBERT \cite{rezayi2022agribert}, BioBERT \cite{lee2020biobert}, and ClinicalRadioBERT \cite{rezayi2022clinicalradiobert}, have demonstrated significant success on domain-specific tasks. Therefore, it is reasonable to speculate that ML-based scoring of students' scientific writing can be improved if we have a domain-specific language model for scientific education. In this case, we need to find the proper domain-specific data that are directly relevant to student writing. It is important to note that student responses are preliminary expressions of general science knowledge and lack the rigor of academic journal publications. In addition, their writing is also influenced by the developmental progress of writing skills and the length of the required tasks. These characteristics of student writing are challenges for using NLP tools to score students' writing \cite{litman2016natural} \cite{ha2016impact}. Therefore, to further improve the application of large pre-trained language models to automatically score students' scientific writing, we use different datasets to train BERT and compare their performance on various downstream tasks. In this work, we make the following contributions: 

1. We provide a method to improve model performance on the downstream tasks by contextualizing BERT with the downstream context in advance. 

2. We prove the effectiveness of domain-specific data in improving BERT-based model performance. 

3. We will release our language models, which can be further tested and used in other science education tasks. 

\section{Methodology}

\subsection{Architecture/Background}
The BERT (Bidirectional Encoder Representations from Transformers) language model \cite{devlin2018bert} is based on the transformer architecture \cite{vaswani2017attention}. It is trained using the masked language modeling (MLM) objective, which requires the model to predict missing words in a sentence given the context. This training process is called pre-training. The pre-training of BERT is unsupervised and only requires unlabeled text data. During pre-training, word embedding vectors are multiplied with three sets of weights (query, key and value) to obtain three matrices \textbf{Q}, \textbf{K}, and \textbf{V}, respectively. These matrices are then used to calculate attention scores, which are weights that measure the importance among input words. For example, in the example "I love my cats.", the word "I" should (ideally) be strongly associated with the word "my", since they refer to the same subject. 

\begin{figure}[ht]
\includegraphics[width=0.5\textwidth]{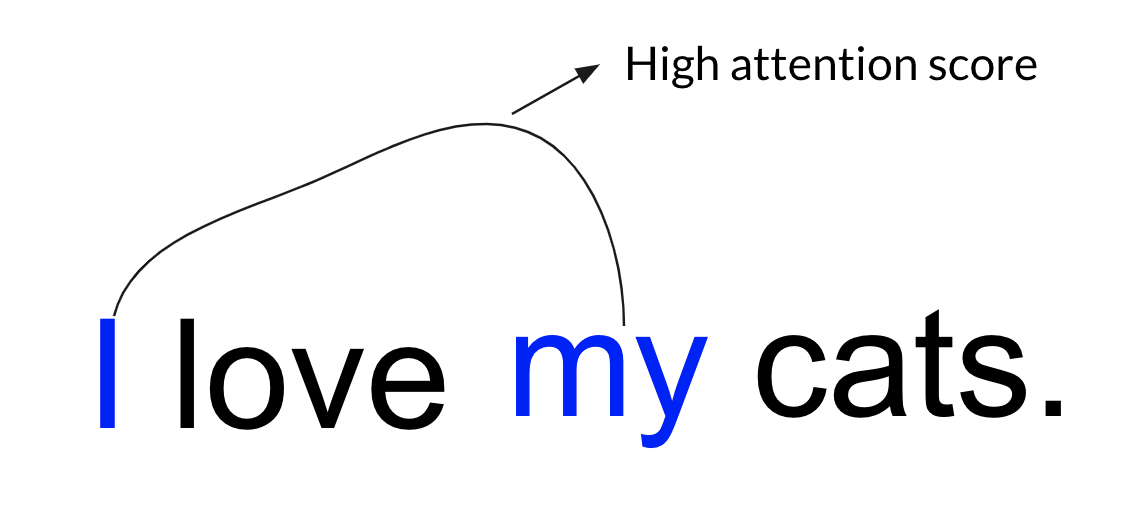}
\caption{An example of BERT's attention mechanism} \label{fig2}
\end{figure}

For each word, the attention scores are then used to weigh intermediate outputs that sum up to the final vector representation of this word.

\begin{equation}
Attention(Q, K, V)=softmax(\frac{QK}{\sqrt{d_k}})V
\end{equation}
where $d_k$ refers to the dimension of the \textbf{K} matrix.

BERT takes a sequence of words as the input, and outputs a latent representation of input tokens in the form of word vectors. This latent representation, or embedding, captures the semantics, positional information, and contextual information of the input sentence. It can be further used for downstream NLP tasks. To use BERT for practical natural language understanding applications, it is necessary to fine-tune the model on the target task. BERT can be fine-tuned on a wide variety of tasks, such as topic classification and question answering, by adding task-specific layers on top of this pre-trained transformer. Fine-tuning is a supervised learning process. During this process, BERT is trained on a labeled dataset and the parameters of the model are updated in training to minimize the task-specific loss function. 

\subsection{Domain-specific training}
BERT is a fundamental building block for language models. In practice, it has many variants that are tailored to the purposes and peculiarities of specific domains \cite{beltagy2019scibert,lee2020biobert,araci2019finbert,rezayi2022agribert,rezayi2022clinicalradiobert}. For example, BioBERT \cite{lee2020biobert} is a large language model trained on biomedical publications (PubMed) and delivers superior performance on biomedical and chemical named entity recognition (NER), since it has a large and contextualized vocabulary of biomedical and chemical terms. 

Substantial evidence indicates that language models perform better when then target and source domains are aligned ~\cite{lee2020biobert,gu2021domain}. In other words, continual pre-training BERT-based models with in-domain corpora could significantly improve their performance on downstream tasks ~\cite{gu2021domain}. In addition, there is much correlation between model performance and the extent of in-domain training. Specifically, training with more relevant in-domain text and training-from-scratch can further improve pre-trained language models \cite{gu2021domain}. 

In this work, we incorporate prior experience in NLP, specifically that of domain-specific training, to train our SCiEdBERT models designed specifically for science education tasks. 

\begin{figure}[ht]
\includegraphics[width=1.2\textwidth]{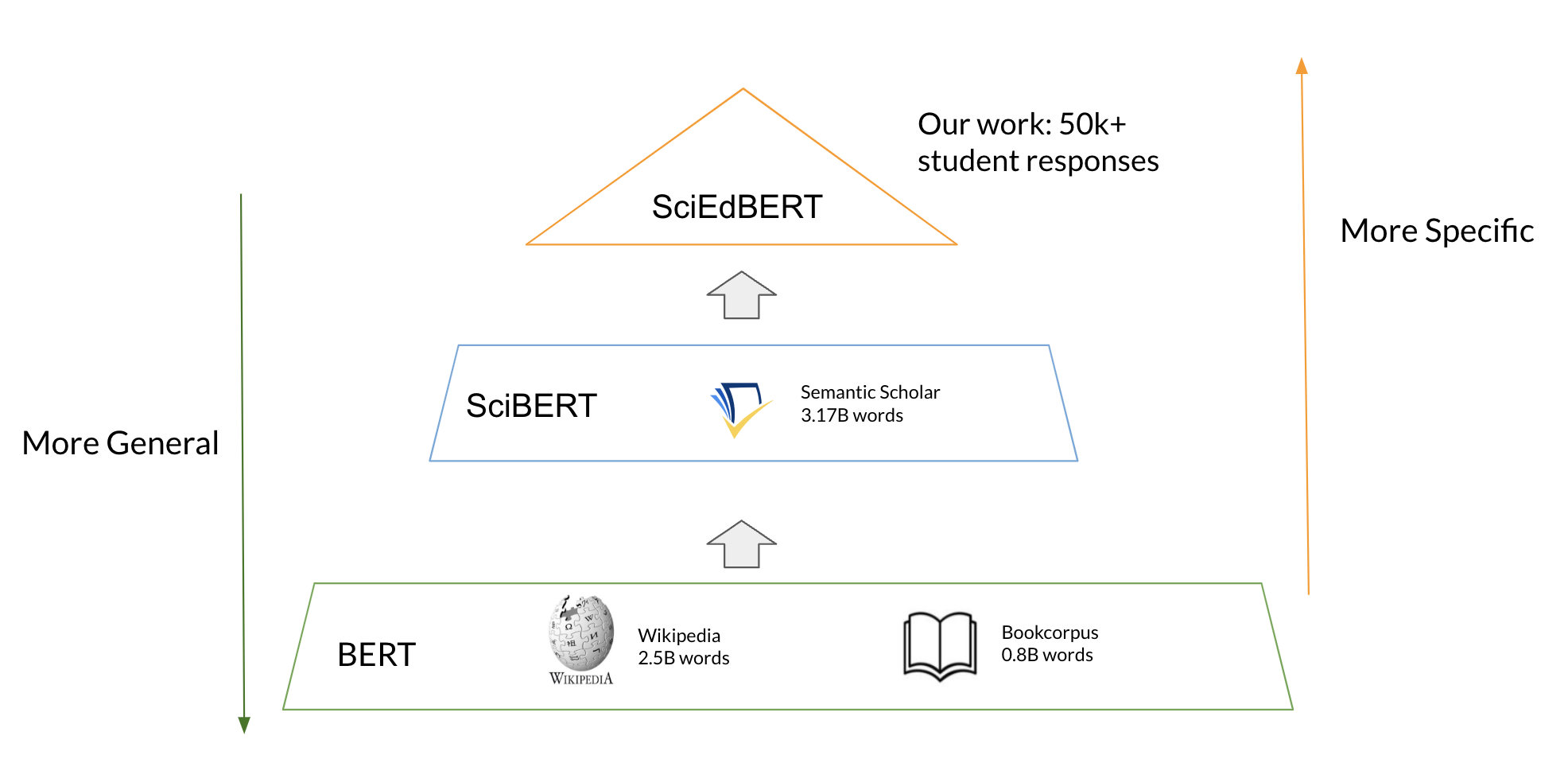}
\caption{The pyramid training scheme} \label{fig3}
\end{figure}

\subsection{Training Design}
We follow a pyramid-shaped training scheme to maximize our models' utilization of domain-relevant data. 

In Figure 3, we can see that SciBERT \cite{beltagy2019scibert} is a science-oriented version of BERT developed through in-domain pre-training. As shown in Table 2, some of the models we developed for this experiment in this study are further extensions of SciBERT through continual pre-training on various science education data. 

The primary benefit of following the pyramid training scheme is to avoid diluting the relatively scarce in-domain data with the vastly more abundant general text data. If instead a language model is trained on a combined corpus of general text and domain-specific data, the effects of in-domain training will be insignificant. 

\section{Experiment}
\subsection{Dataset}
We employ several datasets to train the language models, including the Academic Journal Dataset for Science Education (SciEdJ), a large dataset of students' written Responses (SR1), and a small dataset of students' responses to four argumentation tasks (SR2). Then, we use seven tasks from the large dataset (7T) and the four argumentation tasks (4T) as two datasets to fine-tune the trained language model. Below we briefly introduce these datasets.
\subsubsection{Training Dataset}
We use three datasets to train the language model. The SciEdJ is a collection of 2,000 journal articles from journals in science education. We select ten journals in science education with the highest impact factors according to Web of Science, including \textit{Journal of Research in Science Teaching, International Journal of Science Education, Science Education}, etc. For Each journal, we collect the most recent 200 articles. The SR1 dataset is a collection of over 50,000 student short responses to 49 constructed response assessment tasks in science for middle school students. Students are anonymous to researchers and not traceable. The SR2 dataset is a collection of 2,940 student responses from a set of argumentation assessment tasks~\cite{haudek2021exploring}.

\subsubsection{Fine-tuning Dataset}.
We employ two datasets to evaluate the model performance. The 7T dataset includes seven tasks selected from the SR1 dataset, including short-constructed student responses and human expert-assigned labels. Overall, the 7T dataset includes 5,874 labeled student responses (915 for task H4-2, 915 for task H4-3, 834 for task H5-2, 883 for task J2-2, 743 for task J6-2, 739 for tasks J6-3, and 845 for task R1-2). The 4T dataset includes 2940 student responses and their labels from SR2 dataset (e.g., 770 for item G4, 642 for item G6, 765 for item S2, and 763 for item S3). All the samples in the two datasets are written responses from middle school students to explain science phenomena. Trained experts are employed to assign scores to student responses according to scoring rubrics developed by science education researchers, and the inter-rater reliability is at or above satisfactory level (details see \cite{zhai2022applying}\cite{zhai2022assessing}).

\subsection{Baselines}
Our study aims to examine how the context of training data matters to pre-trained models' (e.g., BERT) performance and explore strategies to further improve model performance. To achieve this goal, we employ various datasets to train and fine-tune the models. First, we use the original BERT as the pre-trained model and 7T as the downstream task. This is the baseline model. We then train a BERT model from SR1 and use 7T as the downstream task. Given that the 7T is grounded in the context of SR1, a comparison between the two fine-tuned models (based on BERT vs. SR1-BERT) can address our goals. 

Second, we repeat this training and fine-tuning process using BERT with SR2 and 4T datasets. To examine the generalization of the findings, we also employ 4T as the downstream task in other pre-trained models, including SciBERT~\cite{beltagy2019scibert}, a BERT model trained on SciEdJ (i.e., SciEJ-BERT), a SciBERT model trained on SciEdJ (i.e., SciEdJ-SciBERT), a BERT model trained on SR2 (i.e., SR2-BERT), and a SciBERT model trained on SR2 (i.e., SR2-SciBERT), with increasingly closer contextualization between the pre-trained models and the downstream tasks.

\subsection{Results}
As Table~\ref{2} presents, the average accuracy of SR1-BERT (0.912) is slightly higher than the accuracy of BERT (0.904). Among the seven tasks, SR1-BERT achieves higher accuracy than BERT on four tasks and are on par with BERT on the remaining three tasks. This indicates that the accuracy of automatic scoring can be improved to a certain extent by training the model with in-domain training data. 

\begin{table*}[]
 \caption{Comparing different model performance on 7T task}
 \label{2}
 \centering
\begin{tabular}{c|cc}
\hline\hline
\multicolumn{1}{c|}{\textbf{Item}} & \multicolumn{2}{c|}{\textbf{Accuracy}} \\
& BERT & SR1-BERT \\
\hline
H4-2 & 0.913 & 0.929\\ 
H4-3 & 0.831 & 0.831\\
H5-2 & 0.958 & 0.970\\
J2-2 & 0.920 & 0.926\\
J6-2 & 0.959 & 0.973\\
J6-3 & 0.845 & 0.845\\
R1-2 & 0.864 & 0.864\\
Average & 0.904 & 0.912 \\
\hline
\hline
\end{tabular}
\end{table*}
This indication is clearer in our second experiment with the 4T dataset. As Table~\ref{1} presents, overall, SR2-SciBERT has the highest average accuracy (0.866), which indicates training the model with the contexts of the downstream tasks can improve the accuracy of automatic scoring.

The model with the second highest accuracy (0.852) is SR2-BERT. SR2-BERT has the same performance as SR2-SciBERT on S3 and even higher accuracy (0.821) than SR2-SciBERT (0.815) on G4. On S2, SR2-BERT's performance (0.915) is only second to SR2-BERT. Only on G6, SR2-BERT has a lower accuracy (0.719) than comparison models. Therefore, although the two models share the same average accuracy, based on the accuracy results on each individual task, SR2-BERT performs better than BERT. This is also in line with our previous findings that context matters in improving model performance.

SciEdJ-SciBERT and SciEdJ-BERT have the lowest average accuracy scores (0.842) among the models. Only on G4 do these two models perform better than BERT. This indicates that the context of science education publications cannot help BERT learn the language of student responses better. In fact, on the contrary, such context may introduce confusion to the machine learning process.


In summary, SR2-SciBERT and SR2-BERT achieve the best results among the models, which indicates that contextualizing the language models with the same language of the downstream tasks can improve the model's performance. 




\begin{table*}[]
 \caption{Comparing model performance on the 4T tasks}
 \label{1}
 \centering
\begin{tabular}{c|ccccc}
\hline\hline
\multicolumn{1}{c|}{\textbf{Item}} & \multicolumn{5}{c|}{\textbf{Accuracy}} \\
& BERT & SciEdJ-BERT & SciEdJ-SciBERT & SR2-BERT & SR2-SciBERT \\
\hline
G4 & 0.792 & 0.804 & 0.815 & 0.821 & 0.815\\ 
G6 & 0.766 & 0.727 & 0.742 & 0.719 & 0.766\\
S2 & 0.895 & 0.882 & 0.889 & 0.915 & 0.928\\
S3 & 0.934 & 0.954 & 0.921 & 0.954 & 0.954\\
Average & 0.847 & 0.842 & 0.842 & 0.852 & 0.866 \\
\hline
\hline
\end{tabular}
\end{table*}

\section{Conclusions}
This study investigates training language models with different contextual data and compares their performance on eleven constructed response tasks. The results indicate that using the in-domain data directly related to downstream tasks to contextualize the language model can improve a pre-trained language model's performance. In automatic scoring of students' constructed responses, this means continual pre-training the language model on student responses and then fine-tuning the model with the scoring tasks. In science education, using SciEdBERT can further improve model performance as SciEdBERT is well-versed in scientific vocabulary. Our study confirms the effectiveness of using domain-specific data to pre-train models to improve their performance on downstream tasks and validate a strategy to adapt language models to science education. 

%
%
%
\newpage
\bibliographystyle{splncs04}
\bibliography{ref}





\end{document}